\documentclass[10pt,twocolumn,letterpaper]{article}

\usepackage{cvpr}      
\usepackage{braket}
\usepackage{graphicx}
\usepackage{amsmath}
\usepackage{amssymb}
\usepackage{booktabs}
\usepackage{multirow}
\usepackage{multicol}
\usepackage{tabularx}
\usepackage{arydshln}
\usepackage{makecell}
\usepackage[T1]{fontenc}

\usepackage[dvipsnames]{xcolor}

\newcommand\blfootnote[1]{%
  \begingroup
  \renewcommand\thefootnote{}\footnote{#1}%
  \addtocounter{footnote}{-1}%
  \endgroup
}

\definecolor{cvprblue}{rgb}{0.21,0.49,0.74}
\usepackage[pagebackref,breaklinks,colorlinks,citecolor=cvprblue]{hyperref}

\def\paperID{14534} 
\def\confName{CVPR}
\def\confYear{2024}

\title{AV2AV: Direct Audio-Visual Speech to Audio-Visual Speech Translation\\with Unified Audio-Visual Speech Representation}

\author{Jeongsoo Choi$^*$\ \quad\quad Se Jin Park$^*$ \quad\quad Minsu Kim$^*$ \quad\quad Yong Man Ro$^\dagger$ \\
School of Electrical Engineering, KAIST, South Korea \\
{\tt\small \{jeongsoo.choi, jinny960812, ms.k, ymro\}@kaist.ac.kr}
}

\begin{document}
\maketitle
\blfootnote{$^*$Equal contribution. $^\dagger$Corresponding author. This work was supported by the National Research Foundation of Korea (NRF) grant funded by the Korea government (MSIT) (No.~NRF-2022R1A2C2005529), Institute of Information \& communications Technology Planning \& Evaluation (IITP) grant funded by the Korea government (MSIT) (No.2022-0-00124, Development of Artificial Intelligence Technology for Self-Improving Competency-Aware Learning Capabilities), and BK21 FOUR (Connected AI Education \& Research Program for Industry and Society Innovation, KAIST EE, No. 4120200113769).}
\begin{abstract}
This paper proposes a novel direct Audio-Visual Speech to Audio-Visual Speech Translation (AV2AV) framework, where the input and output of the system are multimodal (i.e., audio and visual speech). With the proposed AV2AV, two key advantages can be brought: 1) We can perform real-like conversations with individuals worldwide in a virtual meeting by utilizing our own primary languages. In contrast to Speech-to-Speech Translation (A2A), which solely translates between audio modalities, the proposed AV2AV directly translates between audio-visual speech. This capability enhances the dialogue experience by presenting synchronized lip movements along with the translated speech. 2) We can improve the robustness of the spoken language translation system. By employing the complementary information of audio-visual speech, the system can effectively translate spoken language even in the presence of acoustic noise, showcasing robust performance. To mitigate the problem of the absence of a parallel AV2AV translation dataset, we propose to train our spoken language translation system with the audio-only dataset of A2A. This is done by learning unified audio-visual speech representations through self-supervised learning in advance to train the translation system. Moreover, we propose an AV-Renderer that can generate raw audio and video in parallel. It is designed with zero-shot speaker modeling, thus the speaker in source audio-visual speech can be maintained at the target translated audio-visual speech. The effectiveness of AV2AV is evaluated with extensive experiments in a many-to-many language translation setting. Demo page is available on \href{https://choijeongsoo.github.io/av2av}{choijeongsoo.github.io/av2av}.
\end{abstract}    
\begin{figure}[t!]
	\begin{minipage}[b]{\linewidth}
		\centering		\centerline{\includegraphics[width=8.5cm]{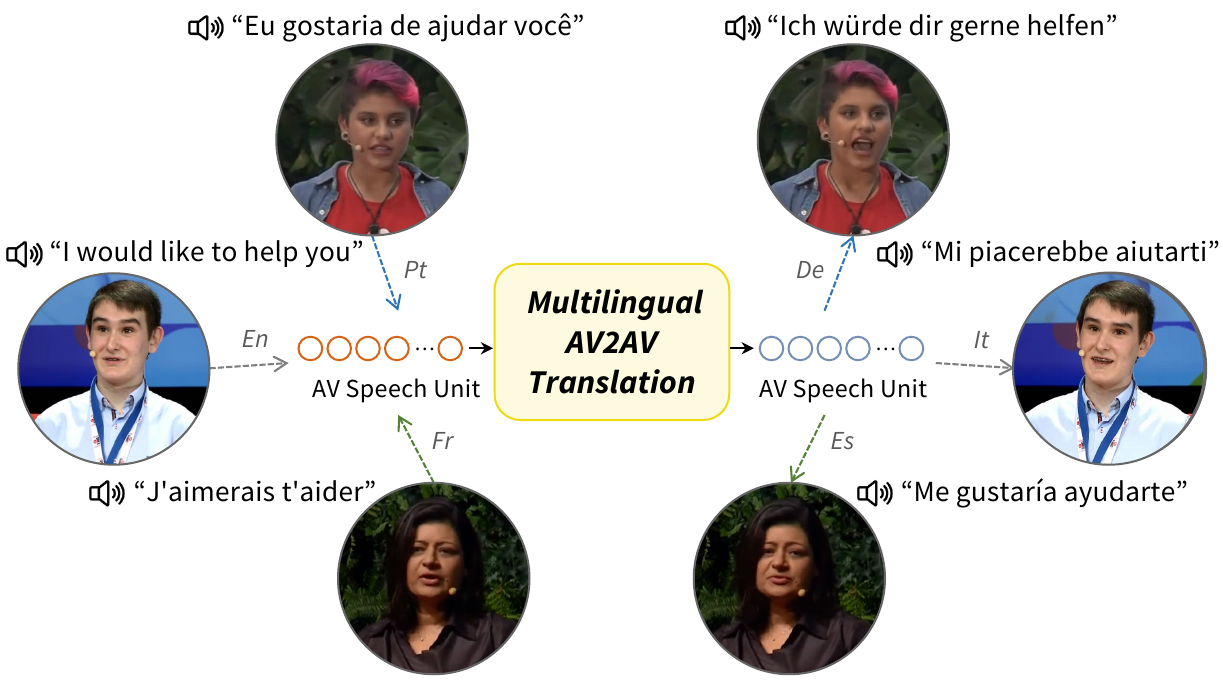}}
	\end{minipage}
	\vspace{-0.5cm}
	\caption{Conceptual illustration of the proposed multilingual Audio-Visual Speech to Audio-Visual Speech Translation (AV2AV) framework. The system can directly translate between multilingual AV speech without requiring any text. Note that the proposed AV2AV can generate both audio speech and visual speech in listener-oriented (\ie, translated) languages.}
	\label{fig:1}
  \vspace{-0.5cm}
\end{figure}

\vspace{-0.7cm}
\section{Introduction}
\label{sec:intro}
In our increasingly interconnected world, where communication transcends linguistic boundaries, Neural Machine Translation (NMT) \cite{sennrich2016improving, johnson2017google, stahlberg2020neural, liu2020multilingual, fan2021beyond, costa2022no} has played a critical role in breaking down barriers in multilingual interaction. Despite their strong performances, NMT systems exhibit limitations in seamless application to virtual conferences or face-to-face interactions. This is due to their reliance on human intervention for text input or speech recognition, as these systems primarily operate with text modalities. Speech-to-Speech Translation (A2A\footnote{Throughout this paper, we employ abbreviations for input and output modalities, using A for Audio, V for Visual, and T for Text.}) \cite{jia2019direct,lee2021direct,lee2021textless,inaguma2022unity,kim2023many} can mitigate this problem by directly translating spoken languages into the target language at the audio level. With the growth of A2A technologies \cite{barrault2023seamlessm4t}, it is anticipated that individuals can effortlessly communicate with one another using their primary languages, irrespective of their nationalities. However, there still exists one unsolved problem in the aspects of multimedia, the discrepancy between the translated speech and the visual stimuli. For example, if the A2A is utilized for video conferencing, people may experience mismatches between what the presented face says and what they hear. This arises because A2A exclusively processes audio speech without addressing the visual component. As inconsistent visual information can negatively affect the perception of speech, which is evidenced by the McGurk effect \cite{mcgurk1976hearing,chung2017lip}, a method jointly considering both audio and visual should be developed.

In this paper, we explore a novel direct Audio-Visual Speech to Audio-Visual Speech Translation (AV2AV) framework. Shown in Fig.~\ref{fig:1}, the proposed AV2AV can directly translate the input Audio-Visual (AV) speech into the desired target language with multimodal experience (\ie, with both audio and visual streams). Therefore, with the proposed AV2AV, 1) We can provide a real face-to-face-like conversation experience enabling participants to engage in discussions using their respective primary languages, mitigating the aforementioned mismatched experience. 2) We can improve the robustness of the system with the complemental information of multimodalities so that the translation can be accurately performed even under noisy environments. 3) We can reduce the computational and maintenance costs, compared to the previous 4-stage cascaded Speech to Audio-Visual Speech Translation (A2AV) approaches \cite{adhikary2023travid,prakash2022technology,waibel2023face,barve2023multi,kr2019towards} which sequentially performed Automatic Speech Recognition (ASR) \cite{amodei2016deep,guo2021recent,kim2022distinguishing}, NMT \cite{pham2020relative,liu2020multilingual}, Text-to-Speech Synthesis (TTS) \cite{ping2017deep,ren2020fastspeech,wu2022adaspeech}, and audio-driven Talking Face Generation (TFG) \cite{prajwal2020lip,park2022synctalkface}. 
In today's world, where millions of multimedia content pieces are generated daily and shared globally in diverse languages, the demand for systems like the proposed AV2AV is anticipated to increase.

However, there is a crucial challenge that must be addressed in the development of a direct AV2AV framework; the lack of translation data between AV speech. Multilingual translation typically requires a large amount of parallel data. While there are relatively abundant text-based datasets \cite{banon2020paracrawl, farhad2021findings, goyal2022flores, ramesh2022samanantar} for NMT and speech-based datasets \cite{anwar2023muavic, wang2020covost, wang2021voxpopuli, iranzo2020europarl} for Speech-to-Text Translation (A2T) \cite{radford2023robust,di2019one,inaguma2019multilingual,li2020multilingual} and A2A \cite{tjandra2019speech,jia2022translatotron,popuri2022enhanced}, there is no publicly available parallel AV2AV corpus.
One possible solution could be generating AV translation data, by separately synthesizing speech and video by applying TTS \cite{wang2017tacotron,lux2022laml,casanova2022yourtts} and TFG \cite{zhang2021flow,zhou2021pcavs,zhang2023sadtalker} onto the existing NMT, A2T, and A2A datasets. Nevertheless, training the model with synthetic face video does not guarantee sufficient final performance, considering the current limitations of TFG in faithfully modeling accurate lip movements \cite{shahzad2022lip,zhou2021joint}.

Instead, our strategy is to employ unified AV speech representations of the recent self-supervised model, AV-HuBERT \cite{shi2022learning}, which is proven to have the modality-agnostic characteristics \cite{hsu2022u,cheng2023mixspeech}. Since its pre-training includes modality dropout, we can reliably obtain unified AV speech representations, whether utilizing audio-only, visual-only, or audio-visual inputs \cite{cheng2023opensr}. Motivated by this, we show that the proposed AV2AV can be trained with audio-only data (\ie, A2A dataset) to perform translation between AV speech. To this end, we introduce a multilingual trained AV-HuBERT (mAV-HuBERT), by pre-training the model with about 7,000 hours of a multilingual AV dataset containing over 100 languages. 
Then, the unified AV speech representations from mAV-HuBERT are discretized through K-means clustering \cite{lakhotia2021generative,lee2021textless,inaguma2022unity}, yielding AV speech units. By treating the discretized AV speech units as pseudo text \cite{kim2023many}, we train our multilingual spoken language translation model. At this time, we employ the modality-agnostic characteristics of AV-HuBERT and utilize audio-only datasets (\ie, A2A datasets) to extract the AV speech units by masking out the visual input stream of mAV-HuBERT. Finally, to render the audio and visual components from the translated AV speech units, we introduce an AV speech unit-based AV-Renderer that can generate synchronized raw speech audio and talking face video in parallel. Especially, the proposed AV-Renderer is designed with zero-shot speaker modeling ability, enabling the preservation of the speaker's voice and face in both audio and video before and after translation.

The contributions of this paper can be summarized as follows:
1) To the best of our knowledge, this is the first work exploring direct Audio-Visual Speech to Audio-Visual Speech Translation (AV2AV), whose inputs and outputs are both audio-visual speech.
2) In order to mitigate the absence of translation data between AV speech, we employ the modality-agnostic characteristics of AV-HuBERT and propose to train the spoken language translation part with the audio-only dataset. At the inference, we show that the trained model with audio-only data can accept any combination of modalities, audio-only, visual-only, and audio-visual inputs, and can produce multimodal speech outputs.
3) For the seamless experience of translated AV speech, we design a zero-shot speaker AV-Renderer. With the proposed AV-Renderer, we can maintain the speaker identity of both audio and video streams before and after translating the AV speech.
4) We explore the AV2AV in a many-to-many language translation setting, so one model can perform X-to-X language translations where X is multilingual, while previous multimodal speech translation systems (\eg, AV2A) can perform only one specific language direction.

\section{Related Works}
\label{sec:relatedworks}

\subsection{Spoken Language Translation}
Neural Machine Translation (NMT) \cite{sennrich2016improving, johnson2017google, stahlberg2020neural, liu2020multilingual, fan2021beyond, costa2022no} has achieved maturity in the modality with the richest data, text. Given that audio modality enables the translation between languages that have no writing systems and the more improved dialogue experience, speech-based translation has emerged. Speech-to-Speech Translation (A2A) \cite{jia2019direct,jia2022translatotron} aims to translate speech from one language to the semantically consistent speech of another language. Early A2A works began with a cascaded approach \cite{lavie1997janus,nakamura2006atr,wahlster2013verbmobil} by sequentially performing ASR, NMT, and TTS. Subsequently, A2T \cite{inaguma2019multilingual,di2019one,li2020multilingual,anwar2023muavic} works were proposed to merge the ASR and NMT stages, and allowed a two-stage A2A system. Recently, it has evolved even into a direct approach \cite{jia2019leveraging, lee2021direct, popuri2022enhanced, huang2022transpeech, kim2023many} that can directly translate speech without relying on intermediate text representation. 

Despite the recent success of audio-based speech translation, multimodal (\ie, audio and visual) speech translation is in its very early stages. One line of research focuses on language translation between audio input and audio-visual output, namely Speech to Audio-Visual Speech Translation (A2AV). They \cite{adhikary2023travid,prakash2022technology,waibel2023face,barve2023multi,kr2019towards} utilized a 4-stage cascaded approach by chaining ASR, NMT, TTS, and TFG. Specifically, audio from the source video is first transcribed into text with an ASR \cite{amodei2016deep}. Then, the text is translated from the source language to the target language through NMT \cite{pham2020relative,liu2020multilingual}. The translated text is synthesized into speech using TTS \cite{ping2017deep,ren2020fastspeech}, and finally, the talking face video is synthesized from the synthesized speech using TFG \cite{prajwal2020lip}. Although the cascaded approach can benefit from the advantage of advancement made in each of the subsystems, they might suffer from slow inference time, domain mismatch, error propagation between the subsystems, and high maintenance costs. 
Most recently, AV-TranSpeech \cite{huang2023av} first proposed direct Audio-Visual Speech-to-Speech Translation (AV2A), whose input is now audio-visual speech and output is audio. By using AV inputs, they showed that the robustness of the translation system on acoustic noise can be improved \cite{afouras2018deep,ma2021end,hong2022visual,hong2023watch}. Also, they tried to solve the insufficient AV2A training data by bringing pre-trained weights for each modal stream from different pre-trained models \cite{shi2022learning,popuri2022enhanced}.

Different from the previous works, this is the first work to explore a direct Audio-Visual Speech to Audio-Visual Speech Translation (AV2AV). The proposed AV2AV incorporates both audio and visual speech as inputs, translates the linguistic content, and produces both audio and visual speech outputs. Moreover, our method is a direct approach where we do not go through with any intermediate text or speech. This is a huge leap from the current 4-stage process in A2AV. Moreover, different from AV-TranSpeech \cite{huang2023av} tried to use different pre-trained models to initialize their model and finetuned it on small size AV2A dataset, the proposed AV2AV model can be trained on an audio-only A2A dataset with unified AV speech representations. The trained model can be directly applied to AV2AV without finetuning. Once trained, we can perform all A2AV, V2AV, and AV2AV, with one single trained model. Finally, we propose a zero-shot AV-Renderer enabling the maintenance of the speaker characteristics of source AV speech at the outputs.

\subsection{Self-supervised Speech Model and Speech Units}
Self-supervised speech models \cite{hsu2021hubert,schneider2019wav2vec,baevski2020wav2vec,chung2021w2v,babu2021xls} have achieved significant performance in various speech processing tasks such as ASR \cite{baevski2020wav2vec, hsu2021hubert}, speaker verification \cite{chen2022wavlm}, and speech translation \cite{babu2021xls,ao2021speecht5,jia2022leveraging}. HuBERT \cite{hsu2021hubert}, one of the prominent self-supervised speech models, is trained to predict hidden units obtained by clustering MFCC features from the masked input, where the hidden units are progressively refined using its learned features. Once the speech features obtained from specific layers are clustered into speech units \cite{chang2023exploration}, they can be utilized as pseudo texts containing the linguistic content of speech. By treating the discrete speech units as pseudo text, various speech processing technologies were proposed \cite{chang2023exploring,maiti2023voxtlm,kim2023towards} such as spoken language modeling \cite{lakhotia2021generative,nguyen2023generative} and speech translation \cite{lee2021direct,lee2021textless,popuri2022enhanced,kim2023many}. 

Extending the modalities, self-supervised multimodal speech models \cite{shi2022learning,haliassos2022jointly,zhu2023vatlm} have been proposed. Among them, AV-HuBERT \cite{shi2022learning} shows promising results in diverse multimodal speech modeling, visual speech recognition \cite{kim2023lip}, AV speech recognition \cite{shi2022robust}, and lip-to-speech synthesis \cite{hsu2023revise,choi2023intelligible}. Recently, its modality-agnostic AV speech representations have drawn big attention in speech recognition \cite{hsu2022u,cheng2023opensr} and Visual Speech Translation (V2T) \cite{cheng2023mixspeech}. Previous works showed that we can robustly get unified AV speech representations with diverse input modalities, audio-only, visual-only, and audio-visual. This is possible because AV-HuBERT proceeds pre-training with modality dropout.

Motivated by the recent success of discrete speech units of HuBERT, we also propose to train our translation system with discretized AV speech units by treating them as pseudo text. As the proposed system is a multimodal system, we employ AV-HuBERT to extract the AV speech units. Moreover, inspired by the modality-agnostic characteristics of AV-HuBERT representations, we propose to train our AV2AV translation model with audio-only datasets (\ie, A2A datasets), by dropping out the visual stream when extracting the AV speech units. To better capture the multilingual speech features, we introduce mAV-HuBERT which is trained on about 7,000 hours of multilingual AV dataset.

\begin{figure*}[t!]
	\begin{minipage}[b]{\linewidth}
	\centering
    \centerline{\includegraphics[width=17cm]{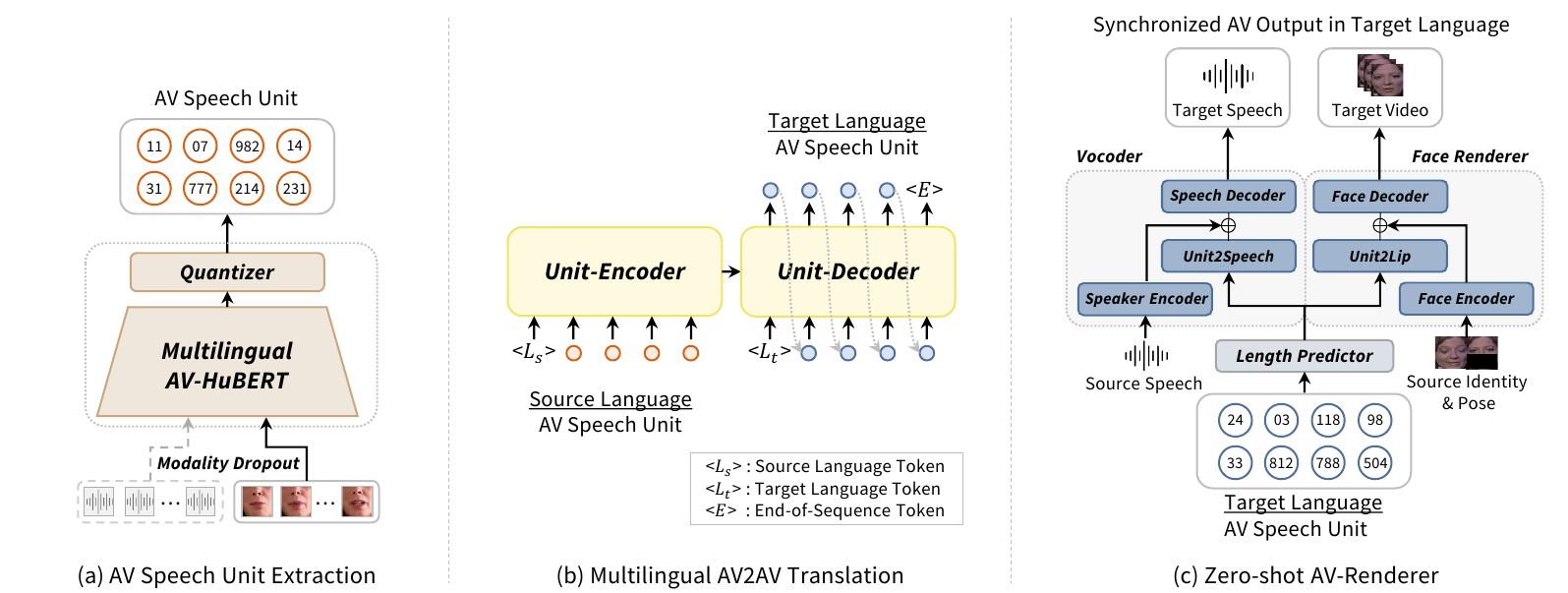}}
	\end{minipage}
	\vspace{-0.5cm}
	\caption{(a) We extract unified audio-visual speech representations using multilingual trained AV-HuBERT. The speech features are discretized into audio-visual speech units through quantization and treated as pseudo text. (b) By using audio-visual speech units, we translate between multilingual languages using a transformer encoder-decoder model. (c) The audio speech and visual speech are generated in parallel from the translated audio-visual speech units by using the proposed Zero-shot AV-Renderer. The renderer can perform in a zero-shot setting so that we can keep the speaker identity the same before and after the translation.}
	\label{fig:2}
  \vspace{-0.5cm}
\end{figure*}

\vspace{-0.5cm}
\section{Proposed Method}
The proposed direct multilingual Audio-Visual Speech to Audio-Visual Speech Translation (AV2AV) framework aims to directly translate both audio stream and visual stream of an input face video from one language to another. To this end, the proposed system is designed with three main parts; 1) extracting linguistic content from the AV input with AV speech units (Fig.~\ref{fig:2}a), 2) performing language translation using the AV speech units (Fig.~\ref{fig:2}b), and 3) synthesizing each modal speech where the linguistic content comes from the translated AV speech units, while the speaker characteristics are controllable (Fig.~\ref{fig:2}c).

\subsection{Unified Audio-Visual Speech Representations}
In order to train the AV2AV system, basically, a parallel corpus of AV speech is required. However, publicly available speech translation datasets are `audio to text' (A2T) \cite{anwar2023muavic}, `audio to audio' (A2A) \cite{wang2021voxpopuli}, and `audio-visual to audio' (AV2A) \cite{huang2023av}. As there is no available `audio-visual to audio-visual' (AV2AV) translation data, it is not feasible to train our model in a parallel AV2AV data setting. To mitigate this, we propose to train our translation model with the audio-only parallel corpus (\ie, A2A dataset), by learning the unified representations of audio and visual speech in advance. Motivated by the recent studies \cite{hsu2022u,cheng2023mixspeech,cheng2023opensr} showing that the AV-HuBERT can extract unified AV speech representations regardless of input modality, our strategy is training our spoken language translation model with the unified AV representations obtained by using audio-only inputs, where the corresponding visual input is dropped out. 

Concretely, we introduce a multilingual trained AV-HuBERT, mAV-HuBERT, to fit it for our multilingual modeling purpose. Different from the English-based AV-HuBERT model \cite{shi2022learning}, mAV-HuBERT is pre-trained on about 7,000 hours of multilingual audio-visual data by combining LRS2 \cite{chung2017lrs}, LRS3 \cite{afouras2018lrs3}, VoxCeleb2 \cite{chung2018voxceleb2}, mTEDx \cite{salesky2021multilingual}, and AVSpeech \cite{ephrat2018looking}, which altogether contains approximately over 100 languages. Then, we discretize the unified AV speech representations of mAV-HuBERT through quantization (\ie, K-means clustering) and obtain AV speech units, following \cite{lakhotia2021generative,kim2023many}. As our mAV-HuBERT is trained with modality dropout similar to AV-HuBERT \cite{shi2022learning}, we can robustly obtain the AV speech units by using different input modalities, audio-only, visual-only, and audio-visual. This enables us to train our AV2AV translation model using an audio-only parallel corpus dataset, while still allowing inference with audio-visual inputs. The processes for obtaining AV speech units are illustrated in Fig.~\ref{fig:2}(a). It is important to note that the discretized speech unit predominantly encompasses speech content, offering a significant advantage as it can be effectively treated as pseudo text \cite{lee2021textless}. Following \cite{lakhotia2021generative,lee2021textless,popuri2022enhanced,inaguma2022unity,kim2023many}, we reduce the length of AV speech units by removing adjacent repeating units.

\subsection{Multilingual Spoken Language Translation}
As shown in Fig.~\ref{fig:2}(b), our AV2AV language translation model is composed of a standard transformer encoder-decoder architecture \cite{vaswani2017attention} which consists of a 12-layer unit-encoder and a 12-layer unit-decoder, similar to a popular NMT system \cite{liu2020multilingual}. Following the recent A2A method \cite{kim2023many} that can perform many-to-many spoken language translation, the unit-encoder takes a source language token <$L_s$> that is for indicating which language should be comprehended and the source AV speech units $\mathbf{u}_s=\{u_s^i\}_{i=1}^{T_s}$, where $T_s$ refers to the length of units. Then, the unit-decoder takes a target language token <$L_t$> that determines the output language and its previous predictions $u_t^{<j}$ to autoregressively predict the next AV speech unit $u_t^j$ of the target language at step $j$. Therefore, the loss function to train the AV2AV language translation can be represented as 
\begin{equation}
\setlength{\abovedisplayskip}{4pt}
\setlength{\belowdisplayskip}{4pt}
\mathcal{L}=-\sum_{j=1}^{T_t}\log p(u_t^j \mid u_t^{<j}, \mathbf{u}_s),
\end{equation}
where $T_t$ refers to the length of target AV speech units.

\begin{figure}[t!]
	\begin{minipage}[b]{\linewidth}
		\centering		\centerline{\includegraphics[width=8.5cm]{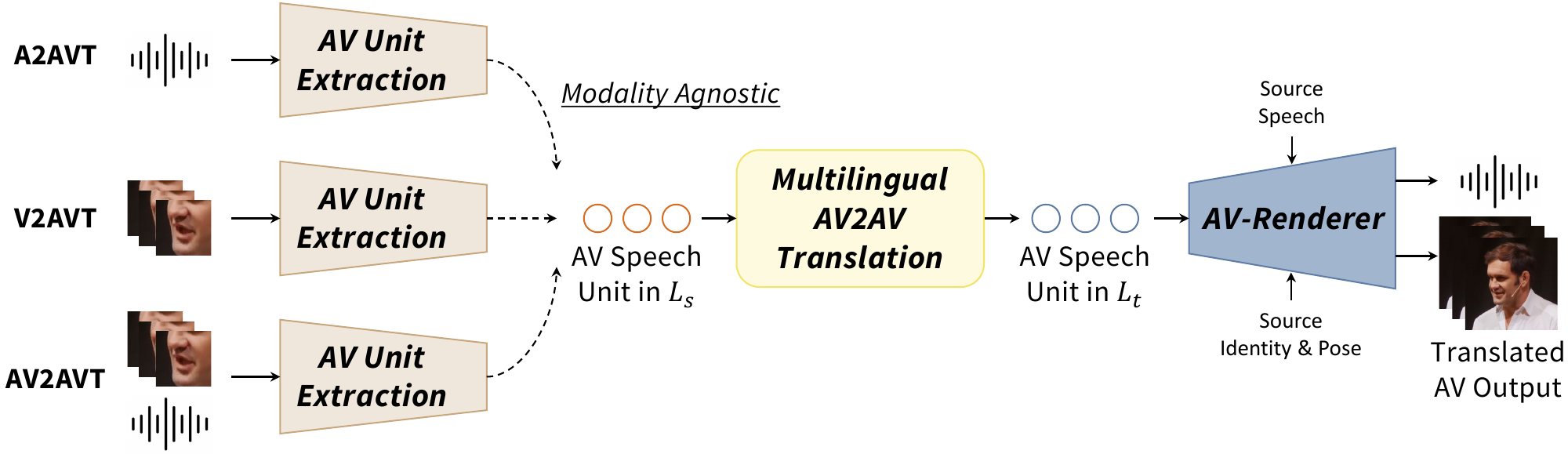}}
	\end{minipage}
	\vspace{-0.5cm}
	\caption{Overview of the proposed AV2AV framework. A single AV2AV model can perform A2AV, V2AV, and AV2AV by learning with unified AV speech representations of mAV-HuBERT.}
	\label{fig:3}
  \vspace{-0.5cm}
\end{figure}

As described before, we can obtain the AV speech units $\mathbf{u}_s$ and $\mathbf{u}_t=\{u_t^i\}_{i=1}^{T_t}$ robustly by using different types of input modalities, audio-only, visual-only, and audio-visual. Therefore, we train our AV2AV language translation model with a large A2A parallel dataset constructed with mTEDx \cite{salesky2021multilingual} and VoxPopuli \cite{wang2021voxpopuli}. The total data amount is about 12,000 hours composed of 19 languages and by reversing the translation direction, the data amount is doubled. Following \cite{kim2023many}, we pre-train the model in many-to-many translation setting on the constructed A2A dataset and finetune the pre-trained model on the target datasets (\ie, MuAViC \cite{anwar2023muavic} and LRS3-T \cite{huang2023av}), respectively. As shown in Fig.~\ref{fig:3}, we show that the model trained with the proposed strategy can perform A2AV, V2AV, and AV2AV with a single model.

\subsection{Zero-shot Audio-Visual Renderer}
The translated AV speech units should be rendered to audio and video to produce the sound and visual movement that humans can perceive. To this end, we introduce a zero-shot AV-Renderer which simultaneously synthesizes the raw audio and video from the AV speech units as shown in Fig.~\ref{fig:2}(c). The crucial aspect is to preserve the speaker identity from the source AV speech to the translated AV speech, ensuring a seamless communication experience for the participants. To achieve this, the proposed AV-Renderer is devised in a zero-shot speaker setting. Only the linguistic content of the translated speech is extracted from the translated AV speech units, while the non-linguistic characteristics are modeled from the source AV speech. The proposed AV-Renderer is comprised of three main components, a length predictor, vocoder, and face renderer.

\textbf{Length Predictor.} As the output audio and visual streams are expected to be synchronized, we only need one common duration modeling for the two output streams. To this end, a length predictor is employed to predict the duration of each AV speech unit. The predicted duration is utilized for synthesizing both audio and visual outputs. 
Similar to that of TTS \cite{ren2020fastspeech}, it consists of two 1D convolution layers with one classifier and is trained with Mean Squared Error loss measured between the prediction and ground truth duration of each unit in the logarithmic domain, as in \cite{lee2021direct,lee2021textless}. The translated AV speech units are repeated by the predicted duration before passing the vocoder and face renderer.

\textbf{Vocoder.} For the vocoder, we basically follow the previous works \cite{lee2021direct,lee2021textless,popuri2022enhanced,kim2023many} that utilized speech unit-based HiFi-GAN \cite{kong2020hifi} and we additionally add the zero-shot speaker modeling ability to the model, as the previous works only support the single-speaker voice. To this end, we leverage a pre-trained speaker verification \cite{jung2019rawnet,desplanques2020ecapa} model of \cite{wan2018generalized} as a speaker encoder to extract the speaker embedding, similar to multi-speaker TTS \cite{jia2018transfer}. It takes a Mel-spectrogram and produces a single speaker embedding, known as a d-vector \cite{variani2014deep}. The d-vector is concatenated to every embedded feature of the AV speech unit where it is embedded by an embedding layer (\ie, Unit2Speech in Fig.~\ref{fig:2}(c)). The concatenated features are fed into the speech decoder whose architecture and training objective are the same as HiFi-GAN \cite{kong2020hifi}, to produce the waveform.

\textbf{Face Renderer.} For the face renderer, we bring the famous audio-driven face synthesis model, Wav2Lip \cite{prajwal2020lip}, and modify it to operate with AV speech units as inputs. As Wav2Lip can generate arbitrary face videos from arbitrary audio, it is appropriate for our zero-shot purpose. Specifically, the source identity feature and pose feature are extracted from a source identity face and a pose prior (upper half of the driving source faces) by a face encoder, as shown in Fig.~\ref{fig:2}(c). The discrete AV speech units are embedded into continuous features through a Unit2Lip encoder which consists of an embedding layer and a transformer layer \cite{vaswani2017attention}. Subsequently, the embedded features and identity features are jointly decoded to generate the target talking face video. It is trained with the same objective functions as \cite{prajwal2020lip}.

\section{Experiment}

\begin{figure*}[t!]
	\begin{minipage}[b]{\linewidth}
	\centering
    \centerline{\includegraphics[width=17.5cm]{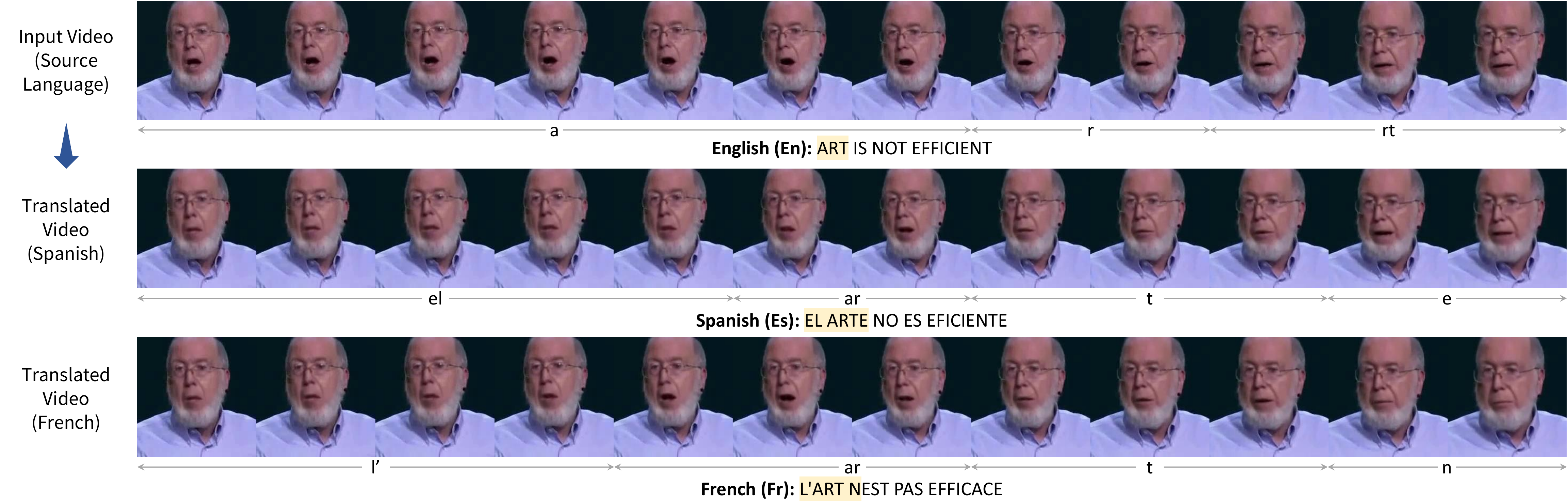}}
	\end{minipage}
	\vspace{-0.6cm}
	\caption{Synthesized outputs of the proposed AV2AV method. The first row is the source input, while the second and third rows are the translated outputs in different target languages. The transcriptions are obtained using an ASR. The highlighted segment of the transcriptions indicates the temporal alignment with the video frames above, and the corresponding phoneme for each frame is marked below. Please note the speaker is maintained in the translated results even if we translate into different languages.}
	\label{fig:4}
  \vspace{-0.5cm}
\end{figure*}

\subsection{Dataset}
For training m-AVHuBERT, we use 7,011 hours of multilingual AV data by combining LRS2 \cite{chung2017lrs}, LRS3 \cite{afouras2018lrs3}, VoxCeleb2 \cite{chung2018voxceleb2}, mTEDx \cite{salesky2021multilingual}, and AVspeech \cite{ephrat2018looking}.

For training AV2AV language translation model, we use about 12k hours of parallel A2A data containing 19 languages constructed with Voxpopuli \cite{wang2021voxpopuli} and mTEDx \cite{salesky2021multilingual} in a many-to-many language translation setting, following \cite{kim2023many}. Then, we finetune the pre-trained model on each evaluation dataset, LRS3-T \cite{huang2023av} and MuAViC \cite{anwar2023muavic}. The LRS3-T \cite{huang2023av} is AV2A data, derived from the LRS3 dataset \cite{chung2017lrs}. It contains the translation direction of En-Es and En-Fr. MuAViC \cite{anwar2023muavic} is an audio-visual speech-to-text translation (AV2T) corpus sourced from TED and TEDx. Since it only provides target text, we generate the target speech by using pre-trained TTS models, VITS \cite{kim2021conditional}, on each language\footnote{Available on \url{https://github.com/coqui-ai/TTS}}. We utilize 4 English (En)-to-X and 4 X-to-En paired data, where X is Spanish (Es), French (Fr), Portuguese (Pt), and Italian (It), which gives a total duration of 2,356 hours. The opposite translation direction is also used for training.

For training the vocoder, we use LRS3 \cite{afouras2018lrs3} for En and mTEDx \cite{salesky2021multilingual} for other languages. For the face renderer, we follow \cite{prajwal2020lip} and utilize the widely used LRS3 \cite{afouras2018lrs3} dataset.

\subsection{Evaluation Metrics}
We evaluate the translation quality and the generation quality of both audio and video. For the translation quality, we employ the BLEU score \cite{papineni2002bleu,post2018call} by transcribing the audio using the off-the-shelf ASR models, following \cite{kim2023many}. To measure the generation quality of video, we adopt metrics used for TFG. This includes Fréchet Inception Distance (FID) \cite{heusel2017gans} to measure visual quality, and LSE-C and LSE-D \cite{prajwal2020lip} to measure the audio-visual synchronization. Also, we conduct a Mean Opinion Score test to measure the naturalness of each respective modality, and the realness of the video, which can be found in Supplementary.

\begin{table}
\renewcommand{\arraystretch}{1.2}
\renewcommand{\tabcolsep}{2.5mm}
\centering
\resizebox{0.9\linewidth}{!}{
\begin{tabular}{clcccc}
\Xhline{3\arrayrulewidth}
\multirow{2}{*}{\textbf{Lang.}} & \multirow{2}{*}{\textbf{Method}} & \multicolumn{4}{c}{\textbf{Finetuning data size (LRS3-T)}} \\ \cmidrule(lr){3-6}
& & 0hr & 10hr & 30hr & 200hr \\ \hline
\multirow{2}{*}{\textbf{En-Es}} & AV-TranSpeech \cite{huang2023av} & - & 21.5 & 22.2 & 45.2 \\
& \textbf{Proposed Method} & \textbf{25.3} & \textbf{39.8} & \textbf{44.6} & \textbf{50.9} \\ \hline
\multirow{2}{*}{\textbf{En-Fr}} & AV-TranSpeech \cite{huang2023av} & - & 24.4 & 28.3 & 33.6 \\
& \textbf{Proposed Method} & \textbf{27.1} & \textbf{34.7} & \textbf{37.2} & \textbf{41.4} \\
\Xhline{3\arrayrulewidth}
\end{tabular}}
\vspace{-0.3cm}
\caption{\label{table:6} AV2A performance comparisons (ASR-BLEU) in low-resource scenario on LRS3-T \cite{huang2023av} dataset. Since the proposed method can be pre-trained on the A2A dataset, it outperforms the previous AV2A model \cite{huang2023av} which relies on the AV2A dataset.}
\vspace{-0.5cm}
\end{table}

\subsection{Implementation Details}
For pre-training mAV-HuBERT, we crop the mouth region into a size of 96$\times$96 by using face detector \cite{deng2020retinaface} and facial landmark detector \cite{bulat2017far}. The cropped mouth frames are converted into grayscale and random cropping and horizontal flipping are applied during training. The audio is resampled to 16kHz and random noise augmentation is applied using MUSAN \cite{snyder2015musan} and overlapped speech, as in \cite{shi2022robust}. We initialize the model with English-trained AV-HuBERT \cite{shi2022learning} and train it with the same training objective with it. To obtain the target cluster, we use pre-trained multilingual HuBERT \cite{hsu2021hubert,lee2021textless} and obtain 1,000 clusters. We train the model for 150k steps with a max token length of 1,000 using 63 GPUs. Our AV speech units are obtained by clustering the unified AV representations into 1,000 clusters.

For training the multilingual AV2AV translation model, we initialize the model with a pre-trained A2A model \cite{kim2023many} and pre-train it on parallel A2A translation data using the AV speech units driven from mAV-HuBERT for 30k steps with a max token length 1,024 using 56 GPUs. The pre-trained model is finetuned on target datasets for 30k steps.

For training the vocoder, we update 1M steps using a batch size of 16 on each language dataset. The face renderer is trained for 35k steps with a batch size of 64. 

\begin{table*}
	\renewcommand{\arraystretch}{1.1}
	\renewcommand{\tabcolsep}{2.0mm}
\centering
\resizebox{0.999\linewidth}{!}{
\begin{tabular}{ccclcccccccc}
\Xhline{3\arrayrulewidth}
\multirow{2}{*}{\textbf{ID}} & \multirow{2}{*}{\makecell{\textbf{Input-Output}\\ \textbf{Modality}}} & \multirow{2}{*}{\makecell{\textbf{Translation} \\ \textbf{System}}} & \multirow{2}{*}{\textbf{Method}} &
\multicolumn{4}{c}{\textbf{X-En}} & 
\multicolumn{4}{c}{\textbf{En-X}} \\ \cmidrule(lr){5-8} \cmidrule(lr){9-12} 

& & &  & \textbf{Es-En} & \textbf{Fr-En} & \textbf{It-En} & \textbf{Pt-En} & \textbf{En-Es} & \textbf{En-Fr} & \textbf{En-It} & \textbf{En-Pt} \\ \hline

\multicolumn{4}{l}{\quad $\bullet$ \textbf{\textit{4-Stage Cascaded System}}} \\ 
{A1} & A-AV & NMT & ASR \cite{anwar2023muavic} + NMT \cite{fan2021beyond} + TTS \cite{xtts} + TFG \cite{prajwal2020lip} &  28.66 & 30.55 & 23.54 & 26.14 & 30.15 & 24.07 & 25.05 & 22.17 \\

{A2} & AV-AV & NMT & AVSR \cite{anwar2023muavic} + NMT \cite{fan2021beyond} + TTS \cite{xtts} + TFG \cite{prajwal2020lip} & 28.70 & 29.21 & 24.54 & 26.30 & 30.09 & 24.56 & 25.41 & 22.66 \\

\hdashline 

\multicolumn{4}{l}{\quad $\bullet$ \textbf{\textit{3-Stage Cascaded System}}} \\ 
{A3} & A-AV & A2T & A2T \cite{anwar2023muavic} + TTS \cite{xtts} + TFG \cite{prajwal2020lip} & 24.06 & 27.01 & 21.92 & 24.11 & 27.81 & 23.42 & 23.29 & 20.46 \\

{A4} & AV-AV & AV2T & AV2T \cite{anwar2023muavic} + TTS \cite{xtts} + TFG \cite{prajwal2020lip} & 24.61  & 26.90 & 22.33 & 24.83 & 28.02 & 23.98 & 23.39 & 20.97 \\

\hdashline 

\multicolumn{4}{l}{\quad $\bullet$ \textbf{\textit{2-Stage Cascaded System (Textless)}}} \\ 
{A5} & A-AV & A2A & A2A \cite{kim2023many} + TFG \cite{prajwal2020lip} &  26.15 & 30.14 & 22.41 & 23.37 & 26.02 & 21.24 & 14.48 & - \\

\hdashline 
\multicolumn{4}{l}{\quad $\bullet$ \textbf{\textit{Direct System (Textless)}}} \\ 
{A6} & A-AV & A2AV & \textbf{Proposed Method} & 26.04 & 31.00 & 22.56 & 24.38 & 27.29 & 23.05 & 16.71 & 12.57 \\
{A7} & V-AV & V2AV & \textbf{Proposed Method} & 10.36 & 6.29 & 8.02 & 5.71 & 18.15 & 15.36 & 11.33 & 9.19 \\
{A8} & AV-AV & AV2AV & \textbf{Proposed Method} & 26.57 & 31.27 & 23.24 & 24.51 & 27.57 & 23.21 & 16.33 & 12.35 \\
\Xhline{3\arrayrulewidth}
\end{tabular}}
\vspace{-0.3cm}
\caption{Translation quality (ASR-BLEU) comparison with baseline systems for X-En and En-X directions on MuAViC. Methods (A1, A2, A3, A4) use the same model for X-En, and 4 different models for En-X directional translations. The methods (A5, A6, A7, A8) use a single model for all 8 directional translations. A2A \cite{kim2023many} method does not have a vocoder to generate Portuguese speech.}
\label{table:1}
\vspace{-0.5cm}
\end{table*}
\begin{table}
	\renewcommand{\arraystretch}{1.2}
	\renewcommand{\tabcolsep}{1.2mm}
\centering
\resizebox{0.999\linewidth}{!}{
\begin{tabular}{llcccccc}
\Xhline{3\arrayrulewidth}
\multirow{2}{*}{\textbf{ID}} & \multirow{2}{*}{\textbf{Method}} & \multirow{2}{*}{\makecell{\textbf{Input}\\\textbf{Modality}}}
& \multicolumn{5}{c}{\centering \textbf{SNR (dB)}} \\ 
\cmidrule(lr){4-8} 
& & & -5 & 0 & 5 & 10 & clean
\\
\hline 
{B1} & {\centering ASR + NMT + TTS + TFG}&  $A$ & $3.73$ & $13.90$  & $20.91$ & $24.84$ & $28.66$ \\
{B2} & {\centering AVSR + NMT + TTS + TFG}&  $AV$  & $15.39$ & $21.93$  & $26.08$ & $26.62$ & $28.70$ \\ \hdashline
{B3} & {\centering A2T + TTS + TFG}& $A$  & $3.02$ & $12.13$  & $18.52$ & $22.75$ & $24.06$ \\
{B4} & {\centering AV2T + TTS + TFG}&  $AV$  & $13.66$ & $18.75$  & $22.04$ & $23.78$ & $24.61$ \\
\hdashline
{B5} & {\centering \textbf{Proposed Method (A2AV)}}& $A$ & $3.64$  & $14.14$ & $19.99$ & $23.87$ & $26.04$ \\
{B6} & {\centering \textbf{Proposed Method (V2AV)}}& $V$ & $10.36$ & $10.36$ & $10.36$ & $10.36$ & $10.36$ \\
{B7} & {\centering \textbf{Proposed Method (AV2AV)}}& $AV$ & $17.31$ & $22.69$ & $23.87$ & $24.20$ & $26.57$ \\
\Xhline{3\arrayrulewidth}
\end{tabular}}
\vspace{-0.3cm}
\caption{\label{table:5} Translation performance (ASR-BLEU) with different input modalities under acoustic noise corruption with different SNR levels (dB) on MuAViC Es-En dataset.}
\vspace{-0.5cm}
\end{table}

\subsection{Baseline Models}
Since there is no previous method that can perform AV2AV, we compare the proposed method with the recently proposed direct AV2A method, AV-TranSpeech \cite{huang2023av} on LRS3-T dataset. Moreover, we compare the translation performance with different cascaded methods on MuAViC \cite{anwar2023muavic} dataset. They are built from state-of-the-art off-the-shelf pre-trained models: AVSR \cite{anwar2023muavic}, ASR \cite{anwar2023muavic}, AV2T \cite{anwar2023muavic}, A2T \cite{anwar2023muavic}, NMT \cite{fan2021beyond}, TTS \cite{xtts}, and TFG \cite{prajwal2020lip}. Please note the objective of the comparisons with the cascaded method is not to achieve state-of-the-art performance, but rather to assess the extent to which the performance of the proposed system can be attained through the direct strategy. 

The AVSR and ASR models \cite{anwar2023muavic} are trained on the multilingual audio-visual corpus, covering 1,200 hours of 9 languages. The AV2T and A2T models \cite{anwar2023muavic} support X-to-En language translations with one model while multiple unidirectional models are used for En-to-X translations. Therefore, we use one pre-trained AV2T or A2T model for \{Es, Fr, It, Pt\}-En translation, while we use 4 individual models for En-Es, En-Fr, En-It, and En-Pt translations. The NMT model \cite{fan2021beyond} is trained on the many-to-many text translation corpus comprising 7.5B sentences for 100 languages. The TTS model \cite{xtts} is a recent state-of-the-art multilingual TTS model by Team Coqui. The TFG model \cite{prajwal2020lip} is one of the popular models that can generate arbitrary identities and languages, trained on 233 hours of audio-visual data.

\subsection{Experimental Results}
\subsubsection{Comparisons with State-of-the-art} 
We compare the AV2A performance with the state-of-the-art direct AV2A method, AV-Transpeech \cite{huang2023av}, by using different amounts of finetuning data, in Table \ref{table:6}. The results show that the proposed method is much more effective than AV-Transpeech, especially in the low-resource setting. As the proposed method can be pre-trained using A2A dataset to perform with different input modalities, the pre-trained model can be directly applied to AV2A without finetuning. In contrast, since AV-Transpeech tries to bring different pre-trained models (\ie, AV-HuBERT \cite{shi2022learning} and A2A \cite{popuri2022enhanced} models) to initialize different parts of their model, the finetuning with the AV2A dataset is mandatory. Even when no finetuning is applied, the proposed method outperforms AV-Transpeech that finetuned on 10hr of AV2A dataset, and achieves comparable performances with the model finetuned on 30hr of dataset. If we finetune the proposed model with the same amount of AV2A dataset as AV-Transpeech, it outperforms AV-Transpeech at all data sizes. This shows the effectiveness of the proposed strategy of learning from A2A dataset by using unified audio-visual speech representations. Fig.~\ref{fig:4} shows the translated results of the proposed AV2AV framework. We show both audio and visual outputs where the audio is transcribed into text through ASR.

\subsubsection{Comparisons with Cascaded Approaches} 
Table~\ref{table:1} shows the BLEU score of different systems for different language pairs. The analysis of results is as follows.

\textbf{(A2, A4 vs. A8)} By comparing the methods that can perform AV2AV, the proposed method (A8) shows comparable results with the text-based translation systems trained on 7.5B parallel data (A2, A4), even though it is trained without using any text supervision. This demonstrates that AV speech units encompass sufficient linguistic content, allowing the training of a multilingual AV2AV translation model solely using the discretized AV speech units\footnote{Moreover, the proposed method has a faster inference time. On average, the proposed AV2AV required 1.36s, whereas the cascaded system (A2) took 2.42s to process 2.03s video (\ie, audio-visual).}.

\textbf{(A1, A2 vs. A3, A4 vs. A5 vs. A6, A8)} We can find that by reducing the cascading stages and shortening the processing time, the BLEU scores become slightly worse overall. This result is attributed to the powerful performance of each advanced subsystem, which benefits from a large-scale dataset. However, we can clearly find that the proposed direct multimodal speech-to-speech translation systems are comparable with the cascaded systems with just a single model. Specifically, the proposed method achieves better performance than the 3-stage cascaded systems (A3, A4) in the X-to-En translation direction. Please note that A1, A2, A3, and A4 use one model for X-to-En and 4 different models to perform En-to-X translation, while the proposed method uses just a single trained model to perform all 8 translation directions in the table. Another important thing is that A5 and the proposed systems are textless so that they can be utilized for the languages that have no writing systems \cite{lee2021textless}, while A1, A2, A3, and A4 cannot be applied to such languages as they rely on text modalities.

\textbf{(A5 vs. A6, A7, A8)} By comparing the proposed method with A2A method (A5), the proposed A2AV method (A6) shows better performances in overall. Moreover, We can confirm that even though the proposed method is pre-trained with the same dataset as A5, the proposed method can be utilized for diverse tasks, including A2AV, V2AV, and AV2AV. This is because the proposed method is trained on unified audio-visual speech representations, eliminating the need for additional training to operate with different modalities. In contrast, A5 utilized audio speech representations during training, so it requires additional finetuning to accept different modal inputs.

\begin{table}
	\renewcommand{\arraystretch}{1.3}
	\renewcommand{\tabcolsep}{4mm}
\centering
\resizebox{0.9\linewidth}{!}{
\begin{tabular}{lcccc}
\Xhline{3\arrayrulewidth}
\multirow{2}{*}{\textbf{Method}} & \multicolumn{4}{c}{\textbf{X-En}} \\\cmidrule(lr){2-5}
& \textbf{Es-En} & \textbf{Fr-En} & \textbf{It-En} & \textbf{Pt-En} \\ \hline
AV-HuBERT \cite{shi2022learning} & 17.63 & 16.83 & 21.74 & 22.22 \\
\textbf{mAV-HuBERT} & \textbf{26.57} & \textbf{31.27} & \textbf{23.24} & \textbf{24.51} \\ \Xhline{3\arrayrulewidth}
\end{tabular}}
\vspace{-0.3cm}
\caption{Ablation study on multilingual translation performance of English-trained AV-HuBERT and multilingual AV-HuBERT.}
\label{table:3}
\vspace{-0.2cm}
\end{table}
\begin{table}
	\renewcommand{\arraystretch}{1.2}
	\renewcommand{\tabcolsep}{1.8mm}
\centering
\resizebox{0.9\linewidth}{!}{
\begin{tabular}{llccccc}
\Xhline{3\arrayrulewidth}
\multicolumn{1}{p{0.5cm}}{\textbf{ID}} & 
\multicolumn{1}{p{3cm}}{\textbf{Method}} & 
\multicolumn{1}{p{1.5cm}}{\centering \textbf{LSE-C} $\uparrow$} & 
\multicolumn{1}{p{1.5cm}}{\centering \textbf{LSE-D} $\downarrow$} & 
\multicolumn{1}{p{1.5cm}}{\centering \textbf{FID} $\downarrow$} \\
\hline
\multicolumn{4}{l}{\quad $\bullet$ \textbf{\textit{Ground Truth}}} \\ 
C1 & GT Audio-Visual & 7.61 & 6.90  & - \\
\hdashline

\multicolumn{4}{l}{\quad $\bullet$ \textbf{\textit{Cascaded System}}} \\ 
C2 & GT Audio + TFG & 8.14  & 6.68  & 3.56 \\
C3 & GT Text + TTS + TFG & 7.36 & 7.06 & 3.56\\

\hdashline
\multicolumn{4}{l}{\quad $\bullet$ \textbf{\textit{Our System (AV-Renderer)}}} \\ 
C4 & GT AV Speech Unit & 7.91  & 6.65 & 3.18 \\
\Xhline{3\arrayrulewidth}
\end{tabular}}
\vspace{-0.3cm}
\caption{Performance comparisons of each renderer on LRS3.}
\label{table:2}
\vspace{-0.5cm}
\end{table} 

\vspace{-0.2cm}
\subsubsection{Analysis of Robustness to Acoustic Noise}
In Table~\ref{table:5}, we evaluate the noise robustness of different systems. Following \cite{shi2022robust}, we randomly perturb the input speech with the sampled babble noise from the test set of \cite{snyder2015musan} with varying SNR levels (-5, 0, 5, 10dB).
By comparing audio-only input systems with multimodal systems (B1, B3, B5 vs. B2, B4, B7), we can confirm the benefits of using AV inputs by obtaining robust performances under acoustic noises. These results show the importance of AV input systems to achieve robust performance in practical situations. The visual input system (B6) is not affected by acoustic noise and even achieves better performance than audio-only systems (B1, B3, B5) under a severely noisy environment (\ie, -5 dB). As one trained AV2AV model can be employed for diverse tasks, we can choose the input modalities appropriate to a given situation.

\vspace{-0.2cm}
\subsubsection{Effectiveness of the mAV-HuBERT}
In order to evaluate the effectiveness of our multilingual AV-HuBERT (mAV-HuBERT) compared to the original English-trained AV-HuBERT of \cite{shi2022learning}, we compare the BLEU scores on MuAViC \cite{anwar2023muavic} X-En translation directions. Table~\ref{table:3} shows the comparison results. The results clearly show that we can improve the multilingual spoken language translation performance largely by pre-training the model with a large-scale multilingual AV dataset (\ie, 7,011 hours with over 100 languages). Conversely, AV-HuBERT trained exclusively in English struggles to accurately capture multilingual speech information, resulting in lower translation performance across all translation directions.

\vspace{-0.2cm}
\subsubsection{Quantitative Evaluation of AV-Renderer}
In Table~\ref{table:2}, we evaluate the visual quality (FID), and degree of synchronization between the generated audio and visual streams (LSE-C and LSE-D). To focus on the performance of renderers, we utilize ground truth audio, text, and AV speech units to generate audio or video for each system. Comparing (C1, C2, C3, C4), the results indicate that the proposed AV-Renderer can generate well-synchronized audio-visual outputs by synthesizing them in parallel, with shared duration information. In contrast, C3 loses accurate synchronization by cascading TTS and TFG, where system error could be propagated between generated samples.

In Table~\ref{table:7}, we evaluate the effectiveness of incorporating zero-shot speaker modeling into the vocoder. To this end, we first compare the BLEU and voice similarity (SIM) using WavLM-TDNN \cite{chen2022wavlm,zhang2023speak} between a single speaker-trained vocoder \cite{kim2023many} and our zero-shot speaker vocoder. The results confirm that by employing speaker embedding, we do not lose the translation performance, and we can improve the voice similarity (SIM). Please note that this is a key component for seamless speech-to-speech translation. To understand the SIM score, we also report that of a popular multi-speaker TTS, YourTTS \cite{casanova2022yourtts}, and we can confirm that the proposed zero-shot speaker vocoder has sufficient ability to maintain the speaker's voice.

\begin{table}
	\renewcommand{\arraystretch}{1.2}
	\renewcommand{\tabcolsep}{2.5mm}
\centering
\resizebox{0.55\linewidth}{!}{
\begin{tabular}{lcc}
\Xhline{3\arrayrulewidth}
\textbf{Vocoder} & \textbf{BLEU} & \textbf{SIM} \\
\hline
YourTTS \cite{casanova2022yourtts} & - & 0.266 \\ \hdashline
Single Speaker & 26.38 & 0.043 \\
\textbf{Zero-Shot Speaker} & 26.57 & 0.222 \\
\Xhline{3\arrayrulewidth}
\end{tabular}}
\vspace{-0.3cm}
\caption{Ablation study using different vocoders on MuAViC Es-En data. To check the performance of our vocoder, we also report the SIM of a popular multi-speaker TTS, YourTTS \cite{casanova2022yourtts}.}
\label{table:7}
\vspace{-0.5cm}
\end{table}

\vspace{-0.1cm}
\section{Conclusion}
\vspace{-0.1cm}
In this paper, we proposed a novel direct Audio-Visual Speech to Audio-Visual Speech Translation (AV2AV) framework. The proposed AV2AV can translate spoken languages in a many-to-many setting without text. By employing multimodal inputs, we can improve the robustness of the system to the acoustic noises. By using multimodal outputs, we can improve the dialogue experience performed in virtual scenarios with no discomfort. The effectiveness of the proposed AV2AV is evaluated with extensive experiments.

{
    \small
    \bibliographystyle{unsrt}
    \bibliography{main}
}

\section{Human Subject Study on Audio-Visual Quality}
In order to evaluate the synthesized quality of audio and video, we conducted a Mean Opinion Score (MOS) test. We gathered 20 participants and asked them to evaluate 32 generated samples of each method to rate in terms of Audio Quality (AQ), Visual Quality (VQ), and overall Realness (R). We presented the audio stream only to evaluate the AQ, the visual stream only for the VQ, and the audio-visual stream altogether to evaluate the R. We compare against the best-performing cascaded system, which is the 4-stage cascaded system of AVSR, NMT, TTS, and TFG. The MOS results are shown in Table~\ref{table:8}. The result demonstrates that we can attain comparable performances with the proposed direct AV2AV approach as with the 4-stage cascaded method comprising state-of-the-art subsystems. Specifically, when both audio and visual streams are simultaneously presented, the participants assess the proposed method more seamlessly generates the two modalities than the cascaded method, as shown in the table (\ie, Realness (R)).

\begin{table}
	\renewcommand{\arraystretch}{1.3}
	\renewcommand{\tabcolsep}{2.0mm}
\centering
\resizebox{0.999\linewidth}{!}{
\begin{tabular}{lcccccc}
\Xhline{3\arrayrulewidth}
& \multicolumn{3}{c}{\textbf{X-En}} & 
\multicolumn{3}{c}{\textbf{En-X}} \\
\cmidrule(lr){2-4} \cmidrule(lr){5-7}
\textbf{Method} & \textbf{AQ} & \textbf{VQ} & \textbf{R} & \textbf{AQ} & \textbf{VQ} & \textbf{R} \\
\hline
AVSR + NMT + TTS + TFG & 3.79 & 4.50 & 3.30 & 3.60 & 3.89 & 3.20   \\
\textbf{Proposed Method (AV2AV)} & 3.97 & 4.60 & 4.00 & 4.51 & 4.11 & 4.07  \\
\Xhline{3\arrayrulewidth}
\end{tabular}}
\vspace{-0.3cm}
\caption{MOS scores on the translated AV output in terms of AQ (Audio Quality), VQ (Visual Quality), and R (Realness).}
\label{table:8}
\vspace{-0.2cm}
\end{table}

\section{Dataset Statistics}
\textbf{For training the m-AVHuBERT}, we use a total of 7,011 hours of the following combined AV datasets. 

\noindent \textbf{LRS2} \cite{chung2017lip} is an English audio-visual dataset, containing 233 hours of training data from British TV shows.

\noindent \textbf{LRS3} \cite{afouras2018deep} is an English audio-visual dataset consisting of approximately 430 hours of video clips from TED and TEDx. 

\noindent \textbf{VoxCeleb2} \cite{chung2018voxceleb2} is a large-scale multilingual corpus for speaker recognition. It has over 1 million utterances from 6,112 celebrities of 145 different nationalities. 

\noindent \textbf{AVSpeech} \cite{ephrat2018looking} is a large-scale multilingual corpus comprising 4700 hours of video clips with no interfering background noises sourced from a total of 290k YouTube videos. 

\noindent \textbf{mTEDx} \cite{salesky2021multilingual} is a multilingual corpus built for speech recognition and speech translation sourced from TEDx talks. It consists of 8 languages; Spanish (Es), French (Fr), Italian (It), Portuguese (Pt), Russian (Ru), Greek (El), Arabic (Ar), and German (De). We use cleaned data of Es, Fr, It, and Pt following \cite{ma2022visual}.

\textbf{For pretraining} the AV2AV language translation model, we use a total of 12k hours of the following A2A datasets. Please refer to \cite{kim2023many} for detailed data statistics for each translation pair. 

\noindent \textbf{Voxpopuli} \cite{wang2021voxpopuli} is a multilingual A2A corpus from European Parliament even recordings. Following \cite{kim2023many}, we use translation from 15 source languages to 15 target languages, which results in 11.2k hours of translation data. 

\noindent \textbf{mTEDx} \cite{salesky2021multilingual} contains speech-to-text translation data from English (En) to Es, Fr, Pt, It, Ru, and El. Since it does not have target speech, we use generated speech from a pretrained TTS model which gives a total duration of 0.7k hours as in \cite{kim2023many}.

\textbf{For evaluation, we finetune} the AV2AV model on the following evaluation datasets. The detailed information for each translation pair is shown in Table~\ref{table:9} and Table~\ref{table:11}

\noindent \textbf{MuAViC} \cite{anwar2023muavic} is a multilingual corpus for audio-visual speech recognition and audio-visual speech-to-text translation (AV2T). It reuses videos of LRS3 and mTEDx datasets including 1,200 hours of transcribed text from over 8000 speakers in 9 languages. Their transcriptions are generated by using an NMT model. Since it only provides target text, we generate the target speech by using pretrained TTS models, VITS \cite{kim2021conditional} on each language. We utilize 4 En-to-X and 4 X-to-En paired data, where X is Es, Fr, Pt, and It, which gives a total of 2,356 hours. 

\noindent \textbf{LRS3-T} \cite{huang2023av} is an AV2A corpus curated from the LRS3 \cite{afouras2018deep} dataset by converting the transcribed English text into the speech in target languages. It results in 200 hours of parallel AV2A translation pairs for En-to-Es and En-to-Fr. 

\begin{table}
	\renewcommand{\arraystretch}{1.3}
	\renewcommand{\tabcolsep}{5mm}
\centering
\resizebox{0.75\linewidth}{!}{
\begin{tabular}{cccc}
\Xhline{3\arrayrulewidth}
\textbf{En-Es} & \textbf{En-Fr} & \textbf{En-It} & \textbf{En-Pt} \\ \hline
437 & 437 & 437 & 437 \\ \hline \hline
\textbf{Es-En} & \textbf{Fr-En} & \textbf{It-En} & \textbf{Pt-En} \\ \hline
178 & 176 & 101 & 153 \\
\Xhline{3\arrayrulewidth}
\end{tabular}}
\vspace{-0.3cm}
\caption{Finetuning dataset amount (hours) of MuAViC for each language pair.}
\label{table:9}
\vspace{-0.5cm}
\end{table}
\begin{table}
	\renewcommand{\arraystretch}{1.3}
	\renewcommand{\tabcolsep}{6mm}
\centering
\resizebox{0.4\linewidth}{!}{
\begin{tabular}{cc}
\Xhline{3\arrayrulewidth}
\textbf{En-Es} & \textbf{En-Fr}  \\ \hline
200 & 200  \\
\Xhline{3\arrayrulewidth}
\end{tabular}}
\vspace{-0.3cm}
\caption{Finetuning dataset amount (hours) of LRS3-T for each language pair.}
\label{table:11}
\vspace{-0.5cm}
\end{table}

\section{Additional Results}
\subsection{Generated Results}
\begin{figure*}[h]
	\begin{minipage}[b]{\linewidth}
	\centering
    \centerline{\includegraphics[width=16cm]{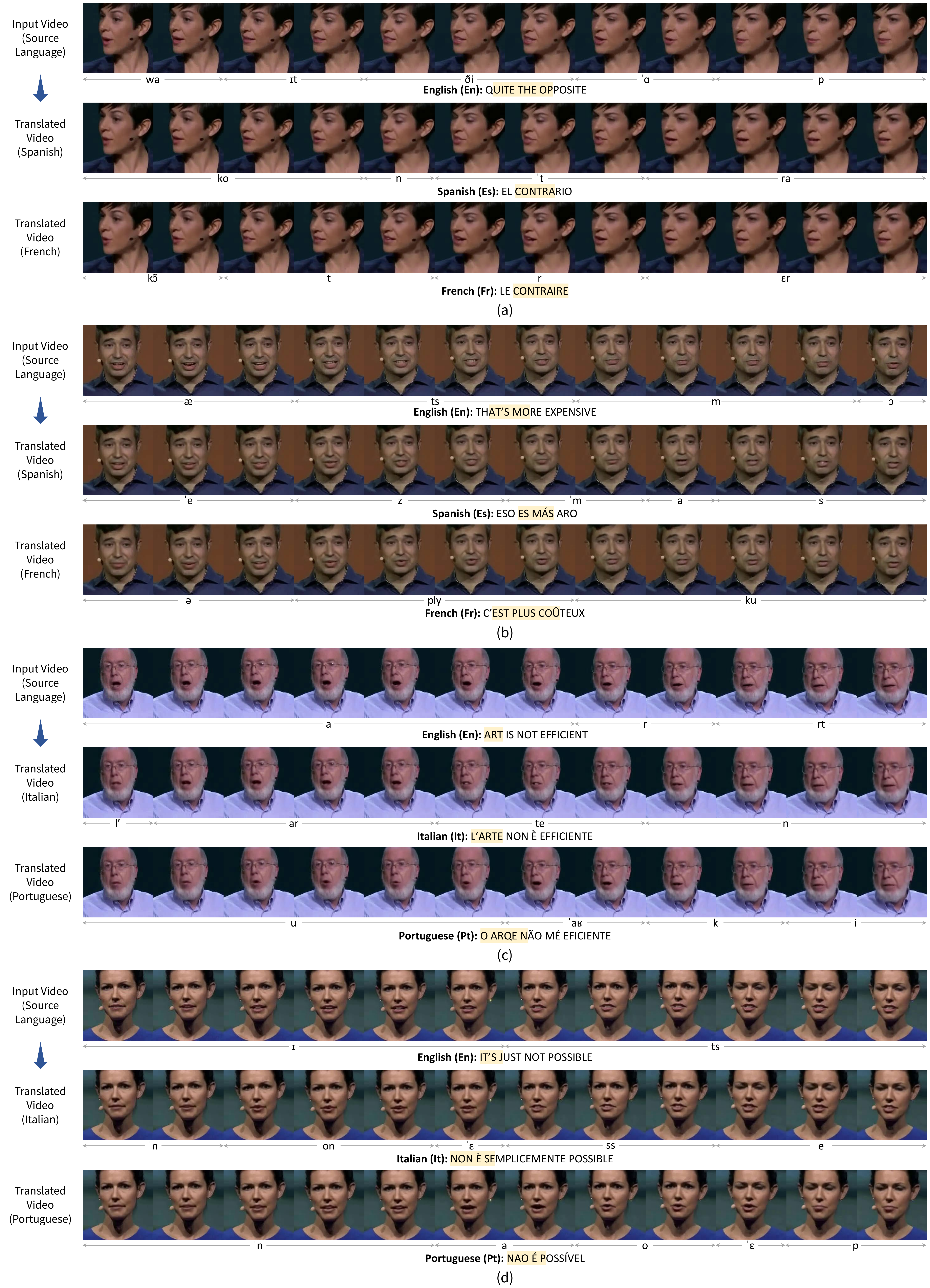}}
	\end{minipage}
	\vspace{-0.5cm}
	\caption{(a-d) Translated results of the proposed AV2AV, each of which the first row is the source input and the second and third rows are the translated outputs in different target languages.}
	\label{fig:6}
  \vspace{-0.5cm}
\end{figure*}

We show more generated translation results of the proposed AV2AV system in Fig. \ref{fig:6}. The AV2AV system seamlessly translates the input video into the target language by transforming the mouth region while keeping the head pose and identity unchanged. The mouth movement aligns well with the corresponding phonemes as written under each video frame. The transcription obtained from ASR of the generated speech is also semantically coherent with the source speech. Moreover, since our model has been trained in many-to-many settings, our model supports translation into multiple different target languages. As shown in the second and third rows of (a-d), it can faithfully generate translated audio and visual speech in different target languages from a single source input. 

Refer to the demo video for more demonstrations of the generated translation results. The demo consists of four parts: the first part shows En-X results on LRS3-T data, the second part shows X-EN results on mTEDx data, the third part compares the proposed method with the best-performing cascaded system, and the last part presents results on a completely different domain, the HDTF \cite{zhang2021flow} data, The HDTF is a high-resolution (720P or 1080P) audio-visual dataset for TFG. For HDTF testing, please note that we have trained the face renderer on the HDTF dataset and attached a face enhancer \cite{wang2021gfpgan} to support high-resolution synthesis of HDTF data. 

\begin{figure}[t!]
	\begin{minipage}[b]{\linewidth}
	\centering
    \centerline{\includegraphics[width=8.5cm]{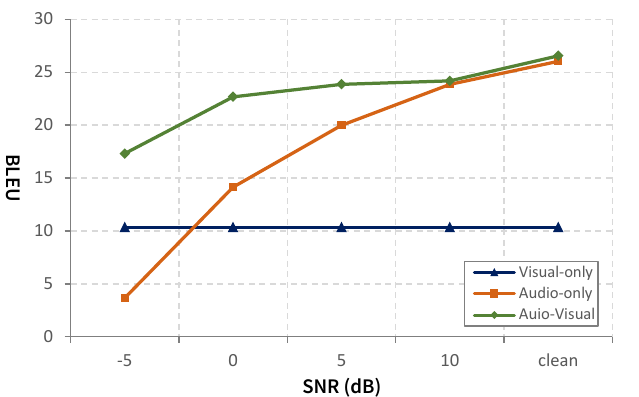}}
	\end{minipage}
	\vspace{-0.5cm}
	\caption{Evaluation of the robustness of the proposed AV2AV system using different input modalities on acoustic noise.}
	\label{fig:5}
  \vspace{-0.5cm}
\end{figure}

\subsection{Noise Robustness}
We visualize the robustness of the proposed AV2AV system to acoustic noise. Specifically, we perturb the clean audio input with the babble noise of MUSAN \cite{snyder2015musan} by varying Signal-to-Noise Ratio (SNR) levels from -5 dB to 10 dB. The BLEU score of the translated results on MuAViC Es-En data is shown in Fig. \ref{fig:5}. As the visual stream is not affected by the acoustic noise, the visual-only case shows the constant BLEU score for all noise levels. In contrast, the audio-only model is mostly affected by the acoustic noise, and the performance is greatly decreased when the noise becomes strong. The interesting thing is that the audio-only model even shows worse performance than the visual-only model in the severely noisy case (\ie, -5 dB). Therefore, it would be better to utilize a visual-only model in a very noisy situation. When we utilize multimodal inputs, audio-visual, the model shows robust performances to the acoustic noise. The model always shows the best performances for all noise levels. Even if the noise is strong, so that the SNR is -5 or 0 dB, the audio-visual model still can robustly translate the speech. The results show the importance of using multimodal inputs for practical usage of speech processing systems.

\begin{table}
	\renewcommand{\arraystretch}{1.3}
	\renewcommand{\tabcolsep}{4.5mm}
\centering
\resizebox{0.7\linewidth}{!}{
\begin{tabular}{cc}
\Xhline{3\arrayrulewidth}
\textbf{Hyper-parameter} & \textbf{Value}  \\
\hline
\# steps & 150k \\
\# warmup steps & 12k \\
LR scheduler & polynomial decay \\
peak learning rate & 5e-4 \\
max frames / GPU & 1000 \\
\# GPUs & 63 \\
Adam ($\beta_1$, $\beta_2$) & (0.9, 0.98) \\
\Xhline{3\arrayrulewidth}
\end{tabular}}
\vspace{-0.3cm}
\caption{Training hyper-parameters of the mAV-HuBERT.}
\label{table:10}
\vspace{-0.2cm}
\end{table}

\begin{table}
	\renewcommand{\arraystretch}{1.3}
	\renewcommand{\tabcolsep}{4.5mm}
\centering
\resizebox{0.7\linewidth}{!}{
\begin{tabular}{cc}
\Xhline{3\arrayrulewidth}
\textbf{Hyper-parameter} & \textbf{Value}  \\
\hline
\# steps & 30k \\
\# warmup steps & 10k \\
LR scheduler & polynomial decay \\
peak learning rate & 3e-4 \\
max tokens / GPU & 1024 \\
\# GPUs & 56 \\
Adam ($\beta_1$, $\beta_2$) & (0.9, 0.98) \\
\Xhline{3\arrayrulewidth}
\end{tabular}}
\vspace{-0.3cm}
\caption{Pretraining hyper-parameters of the AV2AV model.}
\label{table:12}
\vspace{-0.2cm}
\end{table}

\begin{table}
	\renewcommand{\arraystretch}{1.3}
	\renewcommand{\tabcolsep}{4.5mm}
\centering
\resizebox{0.7\linewidth}{!}{
\begin{tabular}{cc}
\Xhline{3\arrayrulewidth}
\textbf{Hyper-parameter} & \textbf{Value}  \\
\hline
\# steps & 30k \\
\# warmup steps & 3k \\
LR scheduler & polynomial decay \\
peak learning rate & 1e-4 \\
max tokens / GPU & 1024 \\
\# GPUs & 64 \\
Adam ($\beta_1$, $\beta_2$) & (0.9, 0.98) \\
\Xhline{3\arrayrulewidth}
\end{tabular}}
\vspace{-0.3cm}
\caption{Fine-tuning hyper-parameters of the AV2AV model.}
\label{table:13}
\vspace{-0.2cm}
\end{table}

\section{Implementation Details}
The mAV-HuBERT has the same architecture as the AV-HuBERT \cite{shi2022learning} large configuration which consists of 24 transformer encoder layers with 16 attention heads, a feed-forward dimension of 4,096, and an embedding dimension of 1,024. We initialize the model with an AV-HuBERT pretrained on 1,759 hours of English subset of LRS3 and VoxCeleb2.  Given the masked audio and visual streams as input, it aims to predict the corresponding target clusters extracted from a pretrained multilingual HuBERT \cite{hsu2021hubert}. During training, we apply modality dropout with a probability of 0.5. To extract the AV speech units, we cluster the unified AV representation into 1,000 clusters using k-mean clustering. Please refer to Table~\ref{table:10} for training configurations. 

The AV2AV model is composed of an encoder embedding layer, 12 transformer encoder layers, 12 transformer decoder layers, and decoder embedding layers. The unit vocabulary size is 1,000 and the embedding dimension of each unit is 1,024. Both the unit encoder and the unit decoder have 8 attention heads and a feed-forward dimension of 4,096. The unit encoder is conditioned on the source language token and the unit decoder generates the target AV speech units conditioned on the target language token. We initialize the model from the mHuBERT unit-based A2A model \cite{kim2023many} and pretrain it based on our AV speech units on parallel A2A translation data for 30k steps with a max token length of 1,024 using the training configuration settings as shown in Table~\ref{table:12}. Then, the pretrained model is further fine-tuned on the evaluation datasets for 30k steps following the configuration settings in Table~\ref{table:13}. The number of GPUs is adjusted according to the dataset size when fine-tuning on the LRS-T dataset and the model with the best performance in the validation set was used. During training, we utilized AV speech units extracted from audio-only, while at inference, we can use AV speech units extracted from any of audio-only, visual-only, and audio-visual data to perform A2AV, V2AV, and AV2AV.

The vocoder is based on the unit-based HiFi-GAN vocoder \cite{kong2020hifi} with an additional speaker encoder \cite{wan2018generalized} to extract the speaker embedding, a d-vector \cite{variani2014deep}. The AV speech units are embedded into 128-dimension through an embedding layer, and the d-vector is also embedded into 128-dimension with a linear projection. The two embeddings from the AVspeech units and the d-vector are channel-wise concatenated at each timestep. We train the vocoder with a length predictor for 1M steps on individual languages with 1 GPU and a batch size of 16. The face renderer is based on a TFG model, Wav2Lip \cite{prajwal2020lip}. The AV speech units are embedded by an embedding layer of dimension 512 and a single transformer Encoder layer, which are channel-wise concatenated with the corresponding identity features. We train the face renderer for 35k steps with 1 GPU and a batch size of 64. Learning rate of 1e-4 and Adam optimizer are used for both the vocoder and the face renderer. 

\section{Advantages of Direct AV2AV Approach}
By employing the proposed direct AV2AV approach, we can gain several advantages compared to utilizing cascaded systems.
\textbf{1) Inference speed.} As the proposed system does not go through with the intermediate text representations, the inference speed is faster than the cascaded system. We compared the inference speed with the best-performed cascaded system (\ie, A2 in Table 2). On average, the proposed AV2AV approach required 1.36s, whereas the cascaded system took 2.42s to process 2.03s video (\ie, audio-visual) for the complete pipeline.
\textbf{2) Model size.} Compared to the cascaded systems, the proposed direct AV2AV approach has a smaller model size. The number of parameters of the cascaded system is 1,490M parameters in total. In contrast, the proposed AV2AV has 732M parameters.
\textbf{3) Text-free ability.} As the proposed AV2AV can be trained without using text data, the system can be served even for languages having no writing systems, while the cascaded systems cannot.

\clearpage

%
\end{document}